\documentclass{article}
\usepackage{kr}

\usepackage{times}
\usepackage{soul}
\usepackage{url}
\usepackage[hidelinks]{hyperref}
\usepackage[utf8]{inputenc}
\usepackage[small]{caption}
\usepackage{graphicx}
\usepackage{amsmath}
\usepackage{amsthm,amssymb}
\usepackage{booktabs}
\usepackage{algorithm}
\usepackage{algorithmic}
\usepackage[dvipsnames]{xcolor}
\usepackage{pifont}

\usepackage{verbatim}
\usepackage[textsize=small, text width=40]{todonotes}
\usepackage{complexity}

\usepackage{tikz}
\usetikzlibrary{arrows.meta, bending}
\usetikzlibrary{matrix, positioning}

\newtheorem{example}{Example}
\newtheorem{theorem}{Theorem}
\newtheorem{proposition}[theorem]{Proposition}

\newtheorem{definition}{Definition}

\newcommand{\cf}{\mathit{cf}}
\newcommand{\stb}{{\mathit{stb}}}
\newcommand{\adm}{\mathit{adm}}

\newcommand{\com}{\mathit{com}}
\newcommand{\grd}{\mathit{grd}}

\newcommand{\labelin}{I}
\newcommand{\labelout}{O}
\newcommand{\labelund}{U}

\newcommand{\lab}{\mathit{Lab}}

\newcommand{\subf}{\mathcal{F}}

\newcommand{\pext}{\mathit{P\textnormal{-}Ext}}
\newcommand{\pacc}{\mathit{P\textnormal{-}Acc}}

\newcommand{\commentout}[1]{}

\title{Advancing Algorithmic Approaches to Probabilistic Argumentation\\ under the Constellation Approach}

\author{%
Andrei Popescu\and
Johannes P. Wallner\\
\affiliations
Institute of Software Technology, Graz University of Technology, Austria
\emails
\{andrei.popescu, wallner\}@ist.tugraz.at
}

\begin{document}

\maketitle

\begin{abstract}
Reasoning with defeasible and conflicting knowledge in an argumentative form is a key research field in computational argumentation. Reasoning under various forms of uncertainty is both a key feature and a challenging barrier for automated argumentative reasoning. It was shown that argumentative reasoning using probabilities faces in general high computational complexity, in particular for the so-called constellation approach. In this paper, we develop an algorithmic approach to overcome this obstacle. We refine existing complexity results and show that two main reasoning tasks, that of computing the probability of a given set being an extension and an argument being acceptable, diverge in their complexity: the former is $\#\P$-complete and the latter is $\#\cdot \NP$-complete when considering their underlying counting problems. We present an algorithm for the complex task of computing the probability of a set of arguments being a complete extension by using dynamic programming operating on tree-decompositions. An experimental evaluation shows promise of our approach. 
\end{abstract}

\section{Introduction}

The field of computational argumentation is nowadays a cornerstone of approaches to automated and rational argumentative reasoning within Artificial Intelligence (AI)~\cite{handbook-vol-1,handbook-vol-2}. Application avenues for this field include legal reasoning~\cite{PrakkenS15}, medical applications~\cite{CyrasOKT21,FoxGGMSP07}, and multi-agent systems~\cite{FanTMW14}, see, e.g., the overview given by~\citeauthor{AtkinsonBGHPRST17}~(\citeyear{AtkinsonBGHPRST17}). 

Central to computational argumentation are formal approaches that define how reasoning is carried out. Common to many forms of argumentative reasoning is the utilization of argumentation frameworks (AFs)~\cite{Dung95}, in which arguments are represented as vertices and directed edges among arguments represent a directed conflict, or attack, relation. Importantly, it oftentimes suffices to abstract the internal structure of arguments, in order to find acceptable (sets of) arguments~\cite{BesnardGHMPST14}.

Argumentation semantics specify which arguments can be deemed acceptable, and several such semantics exist for different purposes~\cite{BaroniCG11}. A prominent property of such semantics is that of admissibility. A set of arguments is admissible if there are no attacks between two arguments in the set (i.e., they are conflict-free), and for each attack from outside the set onto the set there is a counter-attack from within, defending each argument in the set. An admissible set containing all arguments defended is called a complete extension. 

Towards offering advanced forms of argumentative reasoning, AFs have been extended in several directions~\cite{BrewkaPW14}. Recently, approaches that incorporate forms of uncertainty gained traction in research, e.g., by incorporating weights~\cite{DunneHMPW11} or allowing for forms of incompleteness~\cite{BaumeisterNRS18a}. In probabilistic argumentation~\cite{HunterPPRT21,HunterT17}, uncertainty is captured by probabilities of, e.g., arguments and attacks, making them uncertain. Two main approaches to formalization of probabilities in AFs are the epistemic approach~\cite{Hunter13} and the constellation approach~\cite{LiON11}. 

In the constellation approach, subframeworks of a given AF constitute possible scenarios, thereby including only a part of the given AF. Each such subframework is associated with a probability, stating how probable this particular subframework (argumentative scenario) is. Two main reasoning tasks are then to compute the probability that a set is an extension under a specified semantics, such as complete semantics, and (credulous) acceptability of an argument. 
The former is defined as the sum of probabilities of subframeworks where the set is an extension, and the latter as the sum of probabilities of subframeworks where there is an extension containing the specified argument. 

\citeauthor{FazzingaFF19}~(\citeyear{FazzingaFF19}) showed that these two problems are $\FP^{\#\P}$-complete, e.g., for the complete semantics and when the probabilities are compactly represented as marginal probabilities of arguments and attacks with independence assumptions. 
Intuitively, their complexity results indicate that it is very challenging to compute the results: $\#\P$ (or $\FP^{\#\P}$) hard problems are presumed to be very difficult problems to solve. For instance, the archetypical $\#\P$-complete problem is that of counting the number of satisfying truth-value assignments of a Boolean formula~\cite{GomesSS21}. Counting and probabilities are indeed connected, e.g., if each subframework has the same probability, the reasoning tasks above amount to counting the subframeworks satisfying the chosen criteria. 

The high computational complexity was likely so far a major barrier for development of algorithmic approaches and systems for solving tasks in probabilistic argumentation. To the best of our knowledge, there are not many algorithms and systems for the constellation approach for AFs for the computationally hard tasks. As an exception,~\citeauthor{FazzingaFFS19}~(\citeyear{FazzingaFFS19}) provide algorithms for bipolar frameworks~\cite{AmgoudCLL08}, yet for probabilistic AFs their results do not apply to $\FP^{\#\P}$-complete reasoning tasks. 
In this paper we take up this challenge and develop an algorithmic approach to probabilistic argumentation frameworks under the constellation approach. %

Our main contributions are as follows.
\begin{itemize}
 \item We first refine the complexity results by~\citeauthor{FazzingaFF19}~(\citeyear{FazzingaFF19}), who showed that computing the probability of a set being an extension and acceptability share the same complexity. We show, using counting complexity classes, that their underlying counting problems differ in their complexity: 
 the former problem is $\#\P$-complete, for the complete semantics, and the latter is $\#\cdot\NP$-complete for admissible, complete, and stable semantics, indicating a jump in the counting complexity hierarchy. Moreover, even when restricting to acyclic attack structures, the latter problem remains $\#\P$-complete. 
 \item Next we look into algorithms to solve these tasks. Towards efficient reasoning, we give results in preprocessing probabilistic AFs to simplify given instances.  
 \item Inspired by their capability of solving $\#\P$-hard problems, we develop a dynamic programming algorithm for probabilistic AFs utilizing tree-decompositions~\cite{Bodlaender93}. Our algorithm is capable of solving the $\FP^{\#\P}$-complete problem of computing the probability of a set being a complete extension.
 \item We experimentally evaluate a prototype of our algorithm, which will be publicly available under an open license, and show  promise of our approach that solves PAFs up to $750$ arguments, depending on the attack-structure. %
 \item Finally, we discuss extensions of our approach, e.g., to incorporate dependencies among arguments, relaxing independence assumptions. 
\end{itemize}

\noindent
We give further (proof) details in the supplement.

\section{Background}

We recall main definitions of argumentation frameworks (AFs)~\cite{Dung95} and probabilistic argumentation~\cite{HunterPPRT21,HunterT17} under the constellation approach~\cite{LiON11}. Moreover, we recap the notion of tree-width and tree-decompositions~\cite{Bodlaender93} required for our work. 

\begin{definition}
An argumentation framework (AF) is a pair $(A,R)$ where $A$ is a finite set of arguments and $R \subseteq A \times A$ is an attack relation. 
\end{definition}

For a given AF $F=(A,R)$, if $(a,b) \in R$, then $a$ attacks $b$ (in $F$). Similarly, a set $S\subseteq A$ attacks $b \in A$ if there is an $a \in S$ that attacks $b$ (in $F$). 

For specifying possible AFs, we make use of the notion of subframeworks. A subframework of an AF $F=(A,R)$ is an AF $F' = (A',R')$ with $A'\subseteq A$ and $R'\subseteq R$. Note that $F'$ is an AF, thus $R' \subseteq A' \times A'$ holds. 
The set of all subframeworks is denoted by $\subf(F) = \{F' \mid F' \textnormal{ a subframework of } F\}$. 

Argumentation semantics~\cite{Dung95,BaroniCG11} on AFs are defined via functions $\sigma(F)$, for a given AF $F=(A,R)$, that assign subsets of the arguments as $\sigma$-extensions, i.e., $\sigma(F) \subseteq 2^A$. 
Central to AF semantics is the notion of defense.

\begin{definition}
Let $F=(A,R)$ be an AF. 
A set of arguments $S\subseteq A$ defends an argument $a \in A$ if it holds that whenever $(b,a) \in R$ there is a $c\in S$ with $(c,b) \in R$.
\end{definition}

We next define main semantics on AFs. 

\begin{definition}\label{def:semantics}
Let $F=(A,R)$ be an AF.  A set $S\subseteq A$ is 
{\em conflict-free (in $F$)}, 
  if there are no $a, b \in S$, s.t.\ $(a,b) \in R$.
We denote the collection of conflict-free sets of F by $\cf(F)$. 
For a conflict-free set $S \in \cf(F)$, it holds that
\begin{itemize}
\item  
$S\in\stb(F)$
iff $S$ attacks each $a \in A \setminus S$; 
\item 
$S\in\adm(F)$ 
iff  $S$ defends each $a \in S$; 
\item  
$S\in\com(F)$
iff $S\in \adm(F)$ and whenever $a \in A$ is defended by $S$, then $a \in S$; and 
\item 
$S\in\grd(F)$
iff $S\in \com(F)$ and there is no $T \in \com(F)$ with $T \subsetneq S$.
\end{itemize}
\end{definition}

We refer to subsets of arguments that are in $\sigma(F)$ as $\sigma$-extensions and also as an extension under a semantics $\sigma$, for $\sigma \in
\{\stb, \adm, \com, \grd\}$. 

For developing our algorithms later on it will be useful to also consider the labeling-based definitions of the argumentation semantics~\cite{CaminadaG09}. %

\begin{definition}
Let $F=(A,R)$ be an AF. A labeling $L:A \rightarrow \{\labelin,\labelout,\labelund\}$ in $F$ is a function assigning a label to each argument in $F$. 
\end{definition}

Intuitively, $\labelin$ (``in'') signals accepting the argument in a labeling, $\labelout$ (``out'') is interpreted as attacked by accepted arguments, and $\labelund$ (``undecided'') takes neither stance. 
We sometimes use partial labelings, which are labelings where $L$ is partial. 
To distinguish extensions and labelings under a semantics, we use $\sigma$ and $\sigma_{Lab}$, respectively, unless clear from the context. We sometimes view labelings as triples, arguments assigned $\labelin$, $\labelout$, and $\labelund$, respectively.

\begin{definition}
Let $F=(A,R)$ be an AF. 
For a labeling $L$ in $F$ it holds that
\begin{itemize}
 \item $L \in \adm_\lab(F)$ iff for each $a \in A$ we find that 
 \begin{itemize}
  \item $L(a)=\labelin$ implies that $L(b) = \labelout$ if $(b,a) \in R$ and 
  \item $L(a)= \labelout$ implies $\exists (b,a) \in R$ with $L(b) = \labelin$, 
 \end{itemize}
 \item $L \in \com_\lab(F)$ iff $L\in \adm_\lab(F)$ and for $a \in A$ 
 $L(a)=\labelund$ implies $\exists (b,a) \in R$ with $L(b) = \labelund$ and $\nexists (b,a)\in R$ with $L(b) = \labelin$, and 
 \item $L \in \stb_\lab(F)$ iff for all $a \in A$ it holds that $L(a) \neq \labelund$ and $L \in \adm_\lab(F)$. 
\end{itemize}
\end{definition}

There is a direct correspondence between the extension-based and labeling-based view.
For a given AF $F=(A,R)$ and $S\in\cf(F)$, the corresponding labeling $L_S$ is defined for $a \in A$ by 
$L_S(a) = \labelin$ if $a \in S$, $L_S(a) = \labelout$ if $\exists (b,a) \in R \textnormal{ with } b \in S$, and $L_S(a) = \labelund$ if $a \notin S \textnormal{ and } \nexists (b,a) \in R \textnormal{ s.t. } b \in S$.
Conversely, for a given labeling $L$ in $F$, the corresponding extension $S_L$ is defined as $\{a \mid L(a) = \labelin\}$. We recall the main correspondence results~\cite[Theorem 9 and Theorem 10]{CaminadaG09}. 
        
\begin{proposition}
Let $F=(A,R)$ be an AF and $\sigma \in \{\com,\stb\}$. 
\begin{itemize}
 \item 
 If $S \in \sigma(F)$ then $L_S \in \sigma_\lab(F)$ and 
 \item 
 if $L \in \sigma_\lab(F)$ then $S_L \in \sigma(F)$. 
\end{itemize}
\end{proposition}

\begin{figure}
\centering
\begin{tikzpicture}[>=Stealth]
  \tikzset{node style/.style={draw, circle, fill=lightgray!20}}

  \node[node style, dashed] (a) at (0,0.5) {a};
  \node[node style] (b) at (0,-0.5) {b};
  \node[node style, dashed] (c) at (1,-0.5) {c};
  \node[node style] (d) at (1,0.5) {d};
  \node[node style] (e) at (2,0.5) {e};

  \draw[->, dashed] (a) to[bend right] (b);
  \draw[->] (b) to[bend right] (a);
  \draw[->] (a) to[bend left] (d);
  \draw[->] (d) to[bend left] (a);
  \draw[->] (d) to[bend left] (c);
  \draw[->] (c) to[bend left] (d);
  \draw[->] (c) to[bend right] (b);
  \draw[->] (b) to[bend right] (c);
  \draw[->, dashed] (d) to[bend left] (e);
  \draw[->, dashed] (e) to[bend left] (d);

\end{tikzpicture}
\caption{A PAF with certain (solid lines) and uncertain (dashed lines) arguments and attacks. \label{fig:paf-ex}}
\end{figure}
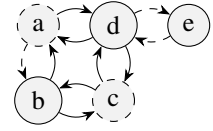

\begin{example}
\label{ex:af}
Consider an AF $F=(A,R)$ with $A = \{a$, $b$, $c$, $d$, $e\}$ and attacks $R= \{(a,b)$, $(a,d)$, $(b,a)$, $(b,c)$, $(c,b)$, $(c,d)$, $(d,c)$, $(d,a)$, $(d,e)$, $(e,d)\}$, see also Figure~\ref{fig:paf-ex} (we explain the difference between solid and dashed components later). 
In this AF, the complete extensions are given by $\com(F) = \{\emptyset,\{e\},\{b\},\{b,e\},\{b,d\},\{a,c,e\}\}$. For instance, set $E=\{a,c,e\}$ is complete since there are no attacks among these arguments (conflict-freeness), all attacks outside are countered (admissibility) and all defended arguments are included. A corresponding complete labeling is $L_E = \{a \mapsto \labelin, b\mapsto \labelout, c \mapsto\labelin, d \mapsto \labelout, e \mapsto \labelin\}$ which is also a stable labeling.
\end{example}

We move on to probabilistic argumentation frameworks under the constellation approach~\cite{LiON11}. 
A probability distribution function (pdf) for an AF $F$ is a function $P: \subf(F) \rightarrow [0,1]$ s.t.\ $\sum_{F' \in \subf(F)} P(F) = 1$. 

\begin{definition}
A probabilistic argumentation framework (PAF) is a triple $(A,R,P)$ where $(A,R)$ is an AF and $P$ is a pdf over $\subf(F)$. 
\end{definition}

For the main part of the paper, we follow the approach dubbed ``IND'', where $P$ is defined via the marginal probabilities of arguments and attacks, by assuming independence, as defined by~\cite{LiON11,Hunter13,FazzingaFF19}. We discuss extensions in Section~\ref{sec:extensions}. 
That is, given a PAF $F=(A,R,P)$, $P: A \cup R \rightarrow [0,1]$ assigns probabilities to arguments and attacks. For each $F'=(A',R') \in \subf(F)$ the probability of the subframework $F'$ is $P(F')$ defined as 
$$\prod_{a \in A'}P(a) \cdot \prod_{a \in A \setminus A'}1-P(a) \cdot 
\prod_{r \in R'}P(r) \cdot \prod_{r \in D \setminus R'}1- P(r)$$
with $D = R \cap (A'\times A')$ the set of all attacks in $R$ between arguments in $A'$. 
That is, the probability of $F'$ is the product of $P(a)$ if $a \in A'$ is present, $1-P(a)$ whenever $a$ was ``removed'' from $A$ (is in $A \setminus A'$), and for each attack $r$, we use $P(r)$ if the attack is present in $F'$ and use $1 - P(r)$ only if the attack was removed, but could be present (i.e., both endpoints are in $A'$). 

By definition, $P(x) = 0$ indicates that subframeworks with $x$ (either argument or attack) have $0$ probability. W.l.o.g., we assume that all arguments and attacks with zero probability are removed from a given PAF $F$. 
For arguments and attacks that are certain (i.e., $P$ assigns probability one), we can restrict our focus to only those subframeworks that contain a certain argument and a certain attack if both endpoints of the attack are present. 
Towards this, we extend the notion of subframeworks via $\subf_P(F) = \{(A',R') \in \mathcal{F}(F) \mid \forall a \in A, P(1)=1 \textnormal{ implies } a \in A' \textnormal{ and } \forall (a,b)\in R, \{a,b\} \subseteq A', P((a,b))=1 \textnormal{ implies} (a,b) \in R' \}$. That is, all subframeworks in $\subf_P(F)$ agree on the certain part (for attacks if both endpoints are there). 

\begin{example}
\label{ex:paf}
Let us continue Example~\ref{ex:af} and include probabilities to the AF $F=(A,R)$ (Figure~\ref{ex:af}). Let $F' = (A,R,P)$ be a PAF, with $P(a)=0.8$, $P(c)=0.9$, $P((a,b))=0.7$, $P((d,e))=0.5$, $P((e,d))=0.3$, and all other probabilities equal to $1$ (solid lines in the figure).
This PAF gives rise to 24 subframeworks in $\subf_P(F')$. We list all subframeworks in the supplement. Consider the subframework $F'' = (\{b, c, d, e\}, \{(b, c), (c, b), (d, c), (c, d), (e,d)\}$. We have  $P(F'') = P(b) \cdot P(c) \cdot P(d) \cdot P(e) \cdot (1 - P(a)) \cdot P((b,c)) \cdot P((c,b)) \cdot P((d,c)) \cdot P((c,d)) \cdot P((e,d)) \cdot (1-P((d,e)) = 1 \cdot 0.9 \cdot 1 \cdot 0.2 \cdot 1 \cdot 1 \cdot 1 \cdot 1 \cdot 0.3 \cdot 0.5 = 0.027 $.
\end{example}

The main reasoning tasks defined on PAFs are computing the probability of a set of arguments being an extension ($\pext$) and the probability of an argument being accepted ($\pacc$)~\cite{FazzingaFF19}. 
\begin{definition}
Let $F=(A,R,P)$ be a PAF, $\sigma \in \{\adm$, $\com$, $\stb\}$ a semantics, and $a \in A$ an argument. 
The probability that $S\subseteq A$ is a $\sigma$-extension (in $F$) is 
$$P^{\mathit{ext}}_\sigma(S) = \sum_{F'\in \subf(F),S \in \sigma(F')}P(F')$$
and the probability of accepting $a$ under $\sigma$ (in $F$) is 
$$P^{\mathit{acc}}_\sigma(a) = \sum_{F'\in \subf(F),\exists S \in \sigma(F'),a \in S}P(F').$$
\end{definition}
In words, the probability of a set of arguments being a $\sigma$-extension is defined as the sum of probabilities of subframeworks for which it holds that they have $S$ as a $\sigma$-extension. Similarly, the probability that $a$ is accepted under $\sigma$ in $F$ is the sum of probabilities of subframeworks where there is a $\sigma$-extension containing the argument. 

\begin{example}
\label{ex:paf-reasoning}
Continuing Example~\ref{ex:paf}, let $S = \{a, c, e\}$, and the query argument be $e$. We obtain $P^{\mathit{ext}}_\com(\{a,c,e\}) = 0.72$ by summing the probabilities of all subframeworks in which $S$ is a complete extension. Similarly we have $P^{\mathit{acc}}_\com(e) = 0.98$ by summing the probabilities of all subframeworks in which there exists a complete extension with argument $e$.
\end{example}

We recall main complexity results. For background on complexity, we refer the reader to standard textbooks on complexity~\cite{Papadimitriou07}. 
Important for probabilistic reasoning are counting complexity classes. A major complexity class for counting is $\#\P$ that contains all functions that count the number of accepting paths of a non-deterministic polynomial-time Turing machine~\cite{Valiant79}. 
\citeauthor{FazzingaFF19}~(\citeyear{FazzingaFF19}) showed that it is $\FP^{\#\P}$-complete to compute the probability of a set of arguments being a complete extension and the problem of computing the probability of an argument being accepted under admissibility, complete and stable semantics, has the same complexity. The complexity of computing the probability of a set of arguments being admissible or stable, on the other hand, can be computed in polynomial time. 

\paragraph{Tree-Decompositions and Tree-Width}

Intuitively, tree-width measures a distance of an undirected graph $G=(V,E)$ to being a tree. A tree-decomposition of a graph $G$ can be, on the one hand, used for measuring tree-width and, on the other hand, is a useful data structure for computation. 
For a given PAF $F=(A,R)$, we associate an undirected graph $F_G=(V,E)$ with $F$ in a direct way: $V= A$ and $E = R$, but we interpret $E$ as undirected edges.

A tree-decomposition of an undirected graph $G=(V,E)$ is a pair $(T,(B_t)_{t\in T})$, with $T$ being a rooted tree and each $B_t \subseteq V $ such that the following properties are satisfied. A $B_t$ is also called a ``bag'' (containing vertices of $G$). 
 Every vertex $x \in A$ is part of a bag $B_t$, i.e., $x \in B_t$ for some $t \in T$.
 For every $(x,y) \in E$ there is a bag $B_t$, $t \in T$, such that $\{x,y\} \subseteq B_t$.
 The set $\{t \mid x \in B_t\}$ induces a subtree of $T$, for each $x \in V$.
In brief terms, a tree-decomposition is a tree such that its nodes (the bags) are connected and all vertices of the original graph $G$ are contained in some bag. The second condition ensures that each edge is present in at least one bag. 
The third condition, often referred to as the connectedness condition, states that whenever two bags $B_t$ and $B_{t'}$ contain a vertex $x$, then on the path between those two bags, we encounter $x$ in all the bags.

The width of a tree-decomposition $(T,(B_t)_{t\in T})$ is the maximum number of vertices in bags minus one, i.e., $\mathit{max}\{\lvert B_t \rvert \mid t \in T\}-1$.
The tree-width of an undirected graph $G=(V,E)$ is the minimum width of all tree-decompositions of $G$. We give illustrations of tree-decompositions in Section~\ref{sec:alg}. 

\section{Complexity Results for Probabilistic AFs}

We investigate the complexity of probabilistic reasoning. In particular, we present novel results for counting variants of probabilistic reasoning. 
\citeauthor{FazzingaFF19}~(\citeyear{FazzingaFF19}) proved the results that, e.g., it is $\FP^{\#\P}$-complete to compute the probability of a set of arguments being a complete extension and to compute the probability of accepting an argument under complete semantics. 

We refine their results, and show that the complexity of problems of computing the probability of a set of arguments and acceptability of an argument diverges on their underlying counting problems. %
Towards our result, we also consider the counting complexity classes $\#\cdot\P$ and $\#\cdot\NP$. 
These ``dot'' classes are from a hierarchy of counting complexity classes~\cite{HemaspaandraV95}, and are defined as follows. 
A counting problem is defined via a witness function $w: \Sigma^* \rightarrow \mathcal{P}^{<\omega}(\Gamma^*)$ that assigns, given a string from alphabet $\Sigma$ a collection of (finite) subsets from alphabet $\Gamma$. The task is to count $|w(x)|$, given $x$. Additionally, we require that each witness $y \in w(x)$ is polynomially bounded by $x$. A counting problem is in $\#\cdot \mathcal{C}$, for a class of decision problems $\mathcal{C}$, if given $x$ and $y$ the problem to decide whether $y \in w(x)$ is in $\mathcal{C}$. For illustration, 
the archetypical $\#\cdot \P$-complete (as for $\#\P$) is $\#\SAT$, with the witness function assigning satisfying truth-value assignments to a formula. 
It holds that $\#\P = \#\cdot\P$, i.e., we can interchangeably use these two classes. 

Let us consider the following counting problems, for a given PAF $F=(A,R,P)$ and semantics $\sigma$:
\begin{itemize}
 \item given a set of arguments $S\subseteq A$, count the number of subframeworks $F' \in \subf_P(F)$ for which we find that $S$ is a $\sigma$-extension in $F'$, and 
 \item given an argument $a \in A$, in how many subframeworks $F' \in \subf_P(F)$ is it the case that there is an $S\in\sigma(F')$ with $a \in S$?
\end{itemize}

\begin{example}
Let $F=(A,R,P)$ be a PAF with all attacks being certain and all arguments having probability $0.5$. By definition, for any $F' \in \subf(F)$ it holds that 
$P(F') = \prod_{a \in A'} 0.5 \cdot \prod_{a \in A \setminus A'} (1-0.5) = 0.5^{|A|}$. 
If there are $n$ many subframeworks for which a specific $S\subseteq A$ is a $\sigma$-extension, then the probability of $S$ being a $\sigma$-extension in $F$ is $n \cdot 0.5^{|A|}$, which is fully determined by $n$. 
\end{example}
As the example suggests, counting the number of subframeworks that satisfy the above criteria acts as special case of computing probabilities of a set being an extension or of the acceptability of an argument.

We start with the problem of counting the number of subframeworks where a given set is an extension. While we will show $\#\P$-completeness ($\#\cdot \P$-completeness), we first argue that this problem is not $\#\P$-hard under parsimonious reductions. 
The type of reductions is important for counting complexity classes, since closure of counting classes under some types of reductions is not immediate~\cite{DurandHK05}. 
A reduction is parsimonious if a counting problem is transformed to another in polynomial time and the number of solutions (cardinality of their respective witness sets) is preserved exactly. 

By a result of~\citeauthor[{{Theorem 2}}]{FazzingaFF20}~(\citeyear{FazzingaFF20}), it holds that one can check in polynomial time whether \emph{some} subframework has a queried set as a complete extension. 
This directly prevents existence of a parsimonious reduction in our case, under complexity theoretic assumptions. If there were a parsimonious reduction from $\#\SAT$ to this problem, one could solve $\SAT$ in polynomial time: one can reduce a Boolean formula to the current problem and in polynomial time find one suitable subframework, which in turn implies that there is a satisfying truth value assignment of the formula.
\begin{proposition}
Unless $\P = \NP$, there is no parsimonious reduction from $\#\SAT$ to counting the number of subframeworks of a given PAF that have a given set of arguments as a complete extension.
\end{proposition}

Nevertheless, we show hardness for a more relaxed notion, Turing reductions, in which a counting problem $A$ is reduced to $B$ by $A$ being solvable in polynomial time, given that we can use $B$ as an oracle. We remark that, as discussed by~\citeauthor{DurandHK05}~(\citeyear{DurandHK05}), for $\#\P$ it is unclear whether closure under Turing reductions holds. 
We conjecture that other suitable types of reductions where closure was proven (like subtractive reductions proposed by~\citeauthor{DurandHK05},~\citeyear{DurandHK05}) can be applied here. In any case, membership in $\#\P$ shows the result for diverging complexity we aim for.

\begin{theorem}
\label{thm:complex-ext}
It is $\#\P$-complete, under Turing reductions, to count the number of subframeworks where a given set of arguments is a complete extension in probabilistic AFs. 
\end{theorem}

For the problem of counting the number of subframeworks where a given argument is accepted, under a semantics, we show $\#\cdot\NP$-completeness, under parsimonious reductions, for the main semantics considered in this work. 

\begin{theorem}
\label{thm:complex-acc}
It is $\#\cdot \NP$-complete, under parsimonious reductions, to count the number of subframeworks where a given argument is accepted in probabilistic AFs under stable, complete, and admissible semantics.
\end{theorem}

In our reduction one part of the constructed PAF is, in fact, acyclic, i.e., the directed graph $(A,R)$ in the PAF is acyclic. This leads to the following result. 

\begin{theorem}
\label{thm:complex-acc-acyclic}
It is $\#\cdot \P$-complete, under parsimonious reductions, to count the number of subframeworks where a given argument is accepted in probabilistic acyclic AFs under grounded, stable, complete, and admissible semantics.
\end{theorem}

That is, even when restricted to an acyclic attack structure, $\#\P$-hardness persists, which is in stark contrast to other abstract argumentation formalisms~\cite{DvorakD18}. 

We next delve into positive complexity results. %
We show %
``fixed-parameter tractability'', under the parameter tree-width. 

An algorithm is a fixed-parameter algorithm~\cite{DowneyF99} w.r.t.\ parameter $k$ for a given problem, 
if the algorithm solves a problem in time $\mathcal{O}(f(k) \cdot n^c)$, with $n$ the size of an instance, $c$ a constant, $k$ a non-negative integer $k$, and $f$ a computable function. 
Intuitively, the running time may depend exponentially on the parameter $k$ and polynomially on the instance size. 
Our algorithm to be presented in Section~\ref{sec:alg} witness fixed-parameter tractability, under the parameter tree-width. In particular, the data structures generated are bounded by the size of bags (i.e., tree-width). 

\begin{theorem}
\label{thm:fpt}
There is a fixed-parameter algorithm, w.r.t.\ the parameter tree-width, for the problem of computing the probability of a set of arguments being a complete extension. %
\end{theorem}

\section{Preprocessing Probabilistic AFs}

Here we look at preprocessing for PAFs. In general, preprocessing is seen as a vital component for solving hard problems, e.g., in SAT solving~\cite{BiereJK21} to simplify instances.

\begin{example}
Consider the PAF shown in Figure~\ref{fig:preproc} with solid components certain and dashed ones uncertain. 
Since argument $a$ is unattacked and certain, all subframeworks in $\subf_P(F)$ contain $a$ and in each complete labeling $a$ is labeled ``in''. For argument $b$, while this argument is attacked by a certain unattacked argument, the attack is uncertain. That is, there is the possibility of having a subframework with $b$ labeled $\labelin$ in a complete labeling (those without the attack $(a,b)$).
On the other hand, argument $c$ can safely by labeled ``out'', whenever $c$ is in a subframework: then also $a$ is present and both attacker and attack are present. 
Argument $d$ is uncertain, but can safely be labeled $\labelin$ in each subframework: in case there is the attacker $c$ and attack $(c,d)$, then $c$ will be labeled $\labelout$. Finally, argument $e$ is attacked by $d$, if $d$ is present, and $d$ is $\labelin$, however, $d$ might not be there at all. Thus there is no fixed stance towards $e$. %
\begin{figure}
\centering
\begin{tikzpicture}[>=Stealth, node distance=2cm]
  \tikzset{node style/.style={draw, circle, fill=lightgray!20}}

  \node[node style, fill=green!15, label=above left:I] (a) at (0,1) {a};
  \node[node style, dashed] (b) at (1.5,1) {b};
  \node[node style, dashed, fill=red!15, label=above left:O] (c) at (1,2) {c};
  \node[node style, dashed, fill=green!15, label=above left:I] (d) at (2.5,2) {d};
  \node[node style] (e) at (4,2) {e};

  \draw[->, dashed] (a) to(b);
  \draw[->] (a) to(c);
  \draw[->, dashed] (c) to (d);
  \draw[->] (d) to (e);

\end{tikzpicture}
\caption{Preprocessing a PAF\label{fig:preproc}}
\end{figure}
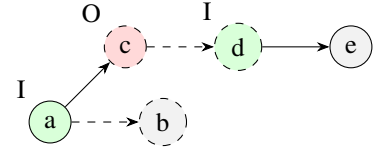
\end{example}

Formally, let us define $\mathcal{A}(F)$, for a given PAF $F=(A,R,P)$ and a given partial labeling $L$, that returns a partial labeling $L'$ as follows. 
 $L'(a)$ is $\labelin$ if for all attacks of $(b,a)$ in $F$ it holds that $L(b) = \labelout$.
 $L'(a)$ is $\labelout$ if there is an attack $(b,a)$ in $F$ we find that $P(b) = 1$, $P(b,a) = 1$ and $L(b) = \labelin$.
We have to label an argument ``in'' if the argument is present in a subframework and for all attacks (if existing) it is clear that they are ``defeated''. An argument is ``out'' if a certain argument labeled $\labelin$ and attack designate it to be out. 
Let $L$ be the least-fixed point of this function. 
It holds that for $\pext$ and a given set $S$, the probability is $0$ if $S$ contains an argument labeled ``out'' in $L$. Moreover, if an argument $a \notin S$ is labeled ``in'' by $L$, then (i) if $a$ is certain, the probability is $0$ and (ii) if $a$ is uncertain only subframeworks without $a$ can have $S$ as a complete extension. 
For $\pacc$ an argument $a$ labeled ``out'' in $L$ has $0$ probability. 

\section{Tree-Decomposition-based Algorithm}
\label{sec:alg}

In this section we develop our algorithm for computing the probabilities of a given set being an extension ($\pext$).
We illustrate our algorithms mainly on the complete semantics, yet our algorithms can be utilized also for admissible sets and stable semantics. Nevertheless, under complete semantics the considered problem is $\FP^{\#\P}$-complete. 

Our algorithm operates on tree-decompositions, more specifically so-called nice tree-decompositions~\cite{BodlaenderK08}, for a given PAF $F=(A,R,P)$.  
A nice tree-decomposition is a tree-decomposition where, additionally, we find that each node is one of the following types:
\begin{itemize}
    \item a root node or a leaf node with empty bags,
    \item a join node $t$ with exactly two children, $t_1$ and $t_2$,  such that $B_t = B_{t_1} = B_{t_2}$, 
    \item an introduction node $t$ with exactly one child $t'$ such that $B_t = B_{t'} \cup \{x\}$ and $x \in A$, or 
    \item a forget node $t$ with exactly one child $t'$ such that $B_t = B_{t'} \setminus \{x\}$ and $x \in A$. 
\end{itemize}
The first condition (empty bags for the root and leaf nodes) is not in the original definition of nice tree-decompositions, but can be imposed directly and is useful for developing our algorithm. A join node has two children and all share the same bag, introduction and forget nodes each have one child and either introduce exactly one or forget exactly one argument. 
Given a (not nice) tree-decomposition, a nice tree-decomposition can be obtained in linear time~\cite{Kloks94}. 

\begin{figure}
\centering
\begin{tikzpicture}[scale = 0.75, node/.style={rectangle, rounded corners, minimum width=1cm, fill=lightgray!20}]

    \node[node] (15) at (0, 0) {$\emptyset$};
    \node[node] (14) at (0,-1) { \{d\}};
    \node[node] (13) at (0,-2) {\{c, d\}};
    \node[node] (12) at (0,-3) {\{a, c, d\}};

    \node[node] (5) at (-1.1,-4) {\{a, c, d\}};
    \node[node] (4) at (-1.1,-5) {\{a, c\}};
    \node[node] (3) at (-1.1,-6) {\{a, b, c\}};
    \node[node] (2) at (-1.1,-7) {\{a, b\}};
    \node[node] (1) at (-1.1,-8) {\{a\}};
    \node[node] (0) at (-1.1,-9) {$\emptyset$};

    \node[node] (11) at (1.1,-4) {\{a, c, d\}};
    \node[node] (10) at (1.1,-5) {\{a, d \}};
    \node[node] (9) at (1.1,-6) {\{d\}};
    \node[node] (8) at (1.1,-7) {\{d, e\}};
    \node[node] (7) at (1.1,-8) {\{d\}};
    \node[node] (6) at (1.1,-9) {$\emptyset$};     

    \draw (0) -- (1);
    \draw (1) -- (2);
    \draw (2) -- (3);
    \draw (3) -- (4);
    \draw (4) -- (5);
    \draw (5) -- (12);
    \draw (6) -- (7);
    \draw (7) -- (8);
    \draw (8) -- (9);
    \draw (9) -- (10);
    \draw (10) -- (11);
    \draw (11) -- (12);
    \draw (12) -- (13);
    \draw (13) -- (14);
    \draw (14) -- (15);

    \node at ([xshift=-2mm]0.west) {0};
    \node at ([xshift=-2mm]1.west) {1};
    \node at ([xshift=-2mm]2.west) {2};
    \node at ([xshift=-2mm]3.west) {3};
    \node at ([xshift=-2mm]4.west) {4};
    \node at ([xshift=-1mm]5.west) {5};
    \node at ([xshift=-2mm]6.west) {6};
    \node at ([xshift=-2mm]7.west) {7};
    \node at ([xshift=-2mm]8.west) {8};
    \node at ([xshift=-2mm]9.west) {9};
    \node at ([xshift=-2mm]10.west) {10};
    \node at ([xshift=-2mm]11.west) {11};
    \node at ([xshift=-2mm]12.west) {12};
    \node at ([xshift=-2mm]13.west) {13};
    \node at ([xshift=-2mm]14.west) {14};
    \node at ([xshift=-2mm]15.west) {15};

    \matrix (table) [matrix of nodes, nodes in empty cells,
        nodes={ minimum height=5mm, anchor= center, scale =0.7, anchor=west}, rounded corners, row sep=-\pgflinewidth, left = 9mm of 13, fill = lightgray!10] (12-table) {
$F:(\{a,c,d\},\{ad,cd,da,dc\}),$ \\ 
$L:(\{a,c\},\{d\},\emptyset), lw:(\{d\},\emptyset), p:0.504$ \\ \hline
$\dots$ \\
    };
    \foreach \rowIndex [count=\rowNumber from 1] in {1, 3} {
        \node[left=1mm of 12-table-\rowIndex-1.west, anchor=east, scale=0.7] { \rowNumber};
    }
    \node[left=10mm of 12-table.west, anchor=south west, scale=0.7] (title) {$\tau_{12}$};

    \matrix (table) [matrix of nodes, nodes in empty cells,
        nodes={ minimum height=5mm, anchor= center, scale =0.7, anchor=west}, rounded corners, row sep=-\pgflinewidth, left = 5mm of 0, fill = lightgray!10] (1-table) {
         $F:(\{a\},\emptyset), L:(\{a\},\emptyset,\emptyset),lw:(\emptyset,\emptyset), p:0.8$ \\
        };
    \foreach \rowIndex [count=\rowNumber from 1] in {1} {
        \node[left=1mm of 1-table-\rowIndex-1.west, anchor=east, scale=0.7] { \rowNumber};
    }
    \node[left=5mm of 1-table.west, anchor=south west, scale=0.7] (title) {$\tau_{1}$};

    \matrix (table) [matrix of nodes, nodes in empty cells,
        nodes={ minimum height=5mm, anchor= center, scale =0.7, anchor=west}, rounded corners, row sep=-\pgflinewidth, left = 4mm of 15, fill = lightgray!10] (13-table) {

        $F\!:\!(\{c,d\},\{cd,dc\}),L\!:\!(\{c\},\{d\},\emptyset),lw\!:\!(\{d\},\emptyset), p\!:\!0.72$
 \\
        };
    \node[above=0mm of 13-table.north west, anchor=south west, scale=0.7] (title) {$\tau_{13}$};

     \matrix (table) [matrix of nodes, nodes in empty cells,
        nodes={ minimum height=5mm, anchor= center, scale =0.7, anchor=west}, rounded corners, row sep=-\pgflinewidth, left = 5mm of 4, fill = lightgray!10] (5-table) {

        $F:(\{a,c,d\},\{ad,da,cd,dc\})$, \\
       $ L:(\{a,c\},\{d\},\emptyset),lw:(\{d\},\emptyset), p:0.72$ \\ \hline
        \dots \\
        };
    \foreach \rowIndex [count=\rowNumber from 1] in {1} {
        \node[left=1mm of 5-table-\rowIndex-1.west, anchor=east, scale=0.7] { \rowNumber};
    }
    \node[left=7mm of 5-table.west, anchor=south west, scale=0.7] (title) {$\tau_{5}$};

    \matrix (table) [matrix of nodes, nodes in empty cells,
        nodes={ minimum height=5mm, anchor= center, scale =0.7, anchor=west}, rounded corners, row sep=-\pgflinewidth, left = 5mm of 2, fill = lightgray!10] (11-table) {

        $F:(\{a,c,d\},\{ad,cd,da,dc\}),$ \\ 
$L:(\{a,c\},\{d\},\emptyset), lw:(\{d\},\emptyset), p:0.504$ \\ \hline

\dots \\
        };
    \foreach \rowIndex [count=\rowNumber from 1] in {1} {
        \node[left=1mm of 11-table-\rowIndex-1.west, anchor=east, scale=0.7] { \rowNumber};
    }
    \node[left=7mm of 11-table.west, anchor=south west, scale=0.7] (title) {$\tau_{11}$};

\end{tikzpicture}
\caption{Nice tree decomposition of the PAF of Example~\ref{ex:paf}\label{fig:paf-td} and corresponding tables (partial) for the computation of $P^{\mathit{ext}}_{com}(\{a, c, e\})$.
} %
\end{figure}
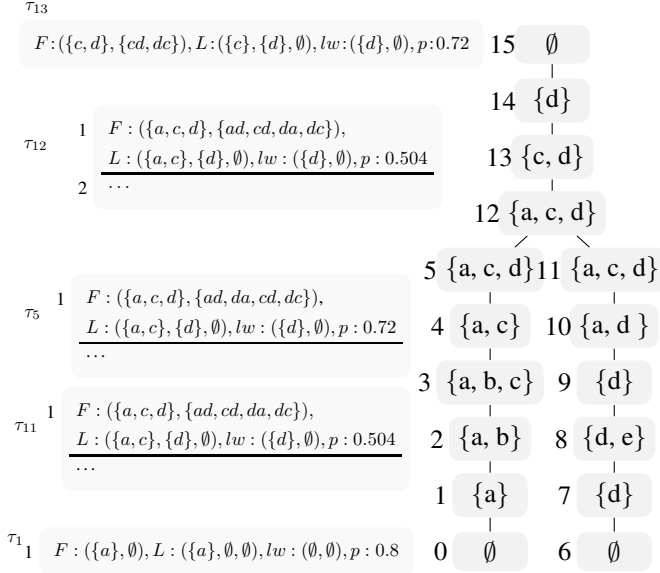

\begin{example}
\label{ex:paf-td}
The PAF from Example~\ref{ex:paf} can be represented by the nice tree-decomposition as shown in Figure~\ref{fig:paf-td}. Here each node has a unique identifier, e.g., the root has the identifier 15. Node 1 introduces argument $a$, node 4 forgets argument $b$, and node 12 is a join node.
\end{example}

For a constant $k$, there is a linear-time algorithm that checks whether a given undirected graph has tree-width $k$, and if reporting positively, also can return a tree-decomposition of minimum width~\cite{Bodlaender96}. Practically, we employ a library~\cite{AbseherMW17} using heuristics for generation for any given PAF. 

The algorithm we develop performs a bottom-up computation on a nice tree-decomposition. Starting from the leaves, for each bag a table of rows is computed. Each row can be seen as a partial solution, applicable to the part of the PAF that was visited so far. %

Algorithm~\ref{alg:main} is the main algorithm, which calls sub algorithms depending on the type of the current node. 
In Line~\ref{alg:main-1} we construct a nice tree-decomposition of the given PAF and initialize an empty table (set) for each bag in Line~\ref{alg:main-2}.
We go over all bags in post-order in the loop in Line~\ref{alg:main-3}. After the case distinction for each type of nodes, we return the probability given in $(\emptyset,\emptyset,p)$ in the root of the tree-decomposition. There is only one such row in each computation. 

\begin{algorithm}
\caption{P-Ext($F=(A,R,P),\sigma,S$)\label{alg:main}}
\begin{algorithmic}[1]

\STATE{Compute nice TD $(T,(B_t)_{t\in T})$ of $(A,R)$} \label{alg:main-1}

\STATE{Let $\tau_t := \emptyset$ for each $t \in T$} \label{alg:main-2}

\STATE{\algorithmicfor\ $t\in T$ in post-order} \label{alg:main-3}

\STATE{\hspace{\algorithmicindent} \algorithmicif\ $t$ is a leaf set $\tau_t := \{(\emptyset,\emptyset,1)\}$ }

\STATE{\hspace{\algorithmicindent} \algorithmicif\ $t$ of type $x \in \{\mathit{Intro},\mathit{Forget}\}$ with child $t'$}

\STATE{\hspace{\algorithmicindent}\hspace{\algorithmicindent} $\tau_t := x(\tau_{t'},B_t,S)$}

\STATE{\hspace{\algorithmicindent} \algorithmicif\ $t$ of type $\mathit{Join}$ with children $t_1,t_2$}

\STATE{\hspace{\algorithmicindent}\hspace{\algorithmicindent} $\tau_t := \mathit{Join}(\tau_{t_1},\tau_{t_2},B_t,S)$}

\STATE{\algorithmicreturn\ $p$ with $(\emptyset,\emptyset,p) \in \tau_t$ with $t$ the root}

\end{algorithmic}
\end{algorithm}

For a given (input) PAF $F = (A,R,P)$, we make use of some auxiliary definitions and shorthands. 
The main datastructure of our algorithms is a row $(s,w,p)$ in a table, which is composed of a structure $s$, a witness $w$, and a probability $p \in [0,1]$. A structure $s = (F',L')$ is a pair of a subframework $F'$ of $\subf_P(F)$ and a labeling $L'$ in $F'$. 
What we call a witness $w$ contains a labeling-witness $lw$, which is a partial labeling in $F'$. %

Intuitively, if a row $(s,w,p)$ is computed in a table $\tau_t$ for node $t$ and bag $B_t$, this means that the sum of probabilities of all ``completions'' of $s$ to arguments in $B_{\leq t}$ that respect the witness $w$ is equal to $p$. 
With $B_{\leq t} = \bigcup_{t' \in T'}B_{t'}$ we define the union of all bags in the subtree $T'$ with root $t$. For instance, $B_{\leq 4}$ of Figure~\ref{fig:paf-td} is $\{a,b,c\}$. A completion of a structure $s$ to $B_{\leq t}$ (all arguments ``seen'') are all completions of the current subframework to subframeworks containing arguments and attacks in $B_{\leq t}$ (and taking certain parts into account), as well as all labelings that extend the one in $s$ up to $B_{\leq t}$. These labelings have to satisfy that those arguments in $S$ that are in $B_{\leq t}$ (i.e., $S\cap B_{\leq t}$) are labeled $\labelin$, and for all arguments already seen but not in the current bag ($B_t \setminus B_{\leq t}$) the conditions of being complete are satisfied. For the arguments inside the bag, whenever the label-witness assigns ``out'' or ``undecided'' to an argument, there must be a reason. %

This means, if one extends the structure $s$ in all possible ways up to $B_{\leq t}$ that still adhere to the conditions of subframeworks of PAFs and complete labelings for those arguments outside the bag, all these structures are ``valid'' subframeworks to sum. 
Since, by construction, the root node has an empty bag, and all arguments have been traversed before, all subframeworks have been considered and their labelings where $S$ is complete (there is exactly one such labeling in each subframework), the result in the root node is the probability of $S$ being a complete extension. 

A witness $w$ is used to ``remember'' facts that decide whether the row can be associated to a complete labeling. 

We next make the above intuition more formal. 
Let $r=(s,w,p)$ be a row in $\tau_t$ for node $t$ with $s=(F',L')$ and $w = lw$, for a given PAF $F=(A,R,P)$. 
Define that $F''=(A'',R'') $ is an expansion of $F'$ and $B_t$ if $A'\subseteq A''$, $R'\subseteq R''$, $(B_t \setminus A')\cap A'' = \emptyset$, and no attacks possible via arguments in $B_t$ that are not in $R'$ are in $R''$. Let $F_{\leq t} = (A_{\leq t},R_{\leq t}) \in \subf_P(F)$ be the subframework s.t.\ $A_{\leq t} = B_{\leq t}$ and $R_{\leq t} = R \cap A_{\leq t}$. Then 
$\subf_P^{\leq t}(F) = \{F'' \in \subf_P(F_{\leq t}) \mid F'' \textnormal{ expands } F' \textnormal{ and }B_t\}$ is the set of all subframeworks with contents restricted to $B_{\leq t}$ expanding $F'$. 

For $L'$, its completion to $A'' \supseteq A'$ is a labeling that assigns the same labels to those in $A'$ as $L'$ and some labeling to those in $A''\setminus A'$. 
A labeling $L''$ on a subframework $F''$ is said to be partially complete w.r.t.\ bag $B_t$, set $S$, and labeling-witness $lw$ if for each argument $a$ in $F''$ the following holds.
If $a \in B_{\leq t} \setminus B_t$ then
    $L(a) = \labelin$ implies all attackers in $F''$ are $\labelout$,
    $L(a) = \labelout$ implies there is an attacker in $F''$ that is $\labelin$, and 
    $L(a) = \labelund$ implies no attacker is $\labelin$ and there is an $\labelund$ attacker.    
For arguments $a \in B_t$, we require that
    $L(a) = \labelin$ implies that all adjacent arguments are $\labelout$, 
    $lw(a) = \labelout$ iff there is an attacker that is $\labelin$ in $F''$, and 
    $lw(a) = \labelund$ iff there is an $\labelund$ attacker und no attacker is $\labelin$.
Finally each argument in $S$ must be assigned ``in''.
That is, the conditions for complete labelings is satisfied for arguments already ``seen'' but forgotten, and for arguments in the current bag, we only require (i) conflict-freeness and (ii) if the labeling-witness specifies an argument to be ``out'' or ``undecided'', there is a justification.

Let $\mathcal{X}_r$ be the set of pairs $(F'',L'')$ such that $F'' \in \subf_P^{\leq t}(F)$ and $L''$ is partially complete in $F''$ w.r.t.\ bag $B_t$, set $S$ and $lw$. It holds that $p$ is the sum of all subframeworks in $\mathcal{X}_r$ (restricted to components in $B_{\leq t}$. Intuitively, $\mathcal{X}_r$ contains all ``compatible'' subframeworks that respect the witness, the current labeling, and the queried set $S$, but relaxing the conditions for the arguments inside the bag, since adjacent arguments might not be traversed yet. %

\paragraph{Introduction Nodes}

Algorithm~\ref{alg:intro} presents our approach to introduction nodes $t$. The child node of $t$ is $t'$ and $\tau'$ contains the rows for $t'$. Intuitively, for each row $r' = (s,w,p)$ from $t'$ we compute new rows $r$, by either adding the introduced argument $a$ or not to the subframework in $s$ (Line~\ref{alg:intro-2} and Line~\ref{alg:intro-5}). If the argument was added, then for each possible way of labeling the new argument (Line~\ref{alg:intro-3}) we compute new labelings and update witnesses. %

Given a PAF $F=(A,R,P)$, as a shorthand, we define $\mathcal{S}_{s+a}(F)$, for a structure $s=(F',L')$ with $F'=(A',R')$, an argument $a$, to be all structures that may extend $s$ with a new argument $a$. 
We define the set of AFs extending $F'$ by adding $a$ and possibly incident attacks. %
\begin{align*}
F'_{+a}=\ &\{F''=(A'',R'') \in \subf_P(F) \mid A'' = A' \cup \{a\}, \\& R' \subseteq R'' \subseteq R' \cup (\{(a,x),(x,a) \mid x \in A''\}\cap R) \}
\end{align*}
Then $\mathcal{S}_{s+a}(F)$ is equal to 
$ \{(F'',L'') \mid F'' \in F'_{+a}, L'' \textnormal{ completion to } F''\}$. %
That is, $\mathcal{S}_{s+a}$ contains all pairs with an AF $F''$ and labeling $L''$ such that $F''$ additionally contains $a$, all arguments and attacks from $F'$, and may add attacks of $a$ (restricted to those originally in $R$). %

When adding an argument $a$, the probability is updated by multiplying with $P(a)$ and considering the factors for each added attack in the structure ($Add(s,a)$) and attack from $R$ not added ($Rem(s,a)$).
\begin{align*}
\mathtt{UpP}(s,a,p) =\ &  p \cdot P(a) \cdot Atts(s,a)\\
Atts(s,a) =\ & \prod_{r \in \mathit{Add}(s,a)} P(r) \cdot \prod_{r \in \mathit{Rem}(s,a)} (1- P(r))
\end{align*}
The witnesses are updated as follows: 
$\mathtt{UpW}(s,w) =\ w \cup \{x \mapsto \labelout \mid (y,x) \in R, L(y)=\labelin\}\ \cup \{x \mapsto \labelund \mid (y,x) \in R, L(y)=\labelund\}$. 
In brief, we store whenever an argument can be assigned to ``out'' or ``undecided'': if there is an attacked argument that is ``undecided'' then its attacker must also be ``undecided'', as required for the complete semantics. An ``in'' attacker is taken care of by requiring in introduction nodes all adjacent arguments to be ``out'', as specified in the following. %

The condition $\sigma_{intro}(s)$ ensures that the labelings in $s$ are conflict-free, i.e., not adjacent arguments are labeled ``in'' and arguments adjacent to ones labeled $\labelin$ are labeled $\labelout$. 
For a resulting set of rows $\tau$, the function $\mathtt{checkAcceptance}(\tau)$ filters those rows where $S$ is not labeled $\labelin$. In practice, this filter can be applied earlier (during row construction). For the sake of readability, we opted to show this filter here. 

\begin{example}
Consider introduction node 1 from Example~\ref{ex:paf-td} (Figure~\ref{fig:paf-td}), and its table $\tau_1$. The unique $r' \in \tau_0$ is the special case $(\emptyset, \emptyset, 1)$.  %
There are two possibilities in terms of subframeworks: adding $a$ or not. Since $a \in S$, the later is filtered out. When adding $a$, we consider new labelings to construct. In this instance, only labeling $a$ to be ``in'' is permitted, since $a \in S$. The probability is then $p = 1 \cdot 0.8 = 0.8$. %
Finally, the labeling-witness stays empty. %
\end{example}

Consider a row $r=(s,w,p)$ created in an introduction node. Then there was previously a row $r'=(s',w',p')$ from which $r$ was constructed. If $r'$ satisfies that $p'$ is the sum of probabilities of subframeworks in $\mathcal{X}_{r'}$ (as defined above), then also $p$ in $r$ satisfies the same property: either $a$ was not added (then it is direct that subframeworks in $\mathcal{X}_r$ are the same as in $\mathcal{X}_{r'}$ and we multiply with $(1-P(a)$) and in the other case, we created a subframework with $a$, some incident attacks, and a labeling that satisfies conflict-freeness. In the latter case, since the bag only increased, the arguments in the bag are not yet fully checked for being complete. %

\begin{algorithm}[b]
\caption{Introduction($\tau',B_t,S$)\label{alg:intro}}
\begin{algorithmic}[1]

\STATE{\algorithmicfor{ $(s,w,p)\in \tau'$}}

\STATE{\hspace{\algorithmicindent}
\algorithmicif{ $P(a) \neq 1$  \algorithmicthen{$\ \tau := \tau \cup \{(s,w,p\cdot (1 - P(a))) \} $} }} \label{alg:intro-2}

\STATE{\hspace{\algorithmicindent} \algorithmicfor{ $s' \in \mathcal{S}_{s+a}(F)$} } \label{alg:intro-3}

\STATE{\hspace{\algorithmicindent}\hspace{\algorithmicindent} \algorithmicif{ $\sigma_{intro}(s')$} \algorithmicthen }

\STATE{\hspace{\algorithmicindent}\hspace{\algorithmicindent}\hspace{\algorithmicindent} $\tau := \tau \cup \{(s',\mathtt{UpW}(w,s'),\mathtt{UpP}(s',a,p))\}$} \label{alg:intro-5}

\STATE{\algorithmicreturn\ $\mathtt{checkAcceptance}(\tau,S)$}

\end{algorithmic}
\end{algorithm}

\paragraph{Forget Nodes}

Let us move to forget nodes (Algorithm~\ref{alg:forget}). The main idea here is that an argument $a$ is ``forgotten'' in a bag (compared to the bag of the unique child in the nice tree-decomposition). This is the point in the algorithm where we can verify that the label assigned to $a$ does not violate the conditions for being complete, since all adjacent arguments were traversed already. The witness-label $lw$ for $a$ enables us to check conditions for $a$.

Say we are in node $t$ with child node $t'$. Consider a row from $t'$ as $r'=(s',w',p')$ with $s' = (F',L')$. If we assume that $p'$ is the sum of probabilities of subframeworks in $\mathcal{X}_{r'}$, then to compute a new row $r = (s,w,p)$, based on $r'$, we need to make sure that for all subframeworks and labels associated with $\mathcal{X}_{r}$ we get that 
(i) the now forgotten argument satisfies the complete conditions and  (ii) that we capture all subframeworks with the new row.

To check the conditions for partial completeness, if $L'(a) = \labelin$, then all attackers in subframeworks in $\mathcal{X}_{r'}$ are labeled ``out'', by assumption of correctness of the child node. If $L'(a) = \labelout$, we need to make sure that there is an ``in'' attacker. This is the case for all subframeworks in $\mathcal{X}_{r'}$ if $lw(a) = \labelout$. Analogously, we verify the partial completeness condition for $L'(a) = \labelund$. 
If one of the checks fails, the row $r'$ is discarded, which we denoted by $\sigma_{forget}(s,w,a)$. By $s-a$ and $w-a$ we denote the removal of $a$ from structure (and incident attacks) and witness.

Finally, after forgetting a node in rows $r'$, multiple rows can coincide on their $(s,w)$ part: these need to be summed (``collapse'' now to one row). 

\begin{example}
Let us consider node 13 which forgets $a$. Here, Algorithm~\ref{alg:forget} first filters out rows $r' \in \tau'$ for which the labeling in $s'$ disagrees with $lw'$ on $a$. A row $r'$ is invalid for further consideration if it labeled $a$ by $O$, while $lw'$ has not witnessed $a$ being attacked by an argument labeled $I$. Analogously for rows where $a$ was labeled undecided. %
\end{example}

\begin{algorithm}[b]
\caption{Forget($\tau',B,S$) \label{alg:forget}}
\begin{algorithmic}[1]

\STATE{$\tau'' := ((s-a,w-a,p) \in \tau' \mid \sigma_{\mathit{forget}}(s,w,a) \textnormal{ is true})$}
\label{alg:forget-1}

\STATE{\algorithmicfor{ $(s,w) \in \{(s',w') \mid (s',w',p')\in \tau''\}$}} %
\label{alg:forget-2}

\STATE{\hspace{\algorithmicindent} $p=\sum_{(s,w,p') \in \tau''}p'$}
\label{alg:forget-3}

\STATE{\hspace{\algorithmicindent} $\tau := \tau \cup \{(s,w,p)\}$ }
\label{alg:forget-4}

\STATE{\algorithmicreturn\ $\tau$}

\end{algorithmic}
\end{algorithm}

\paragraph{Join Nodes}

The idea of join nodes $t$ is to ``merge'' tables from two children nodes $t_1$, $t_2$, with different $B_{\leq t_1}$ and $B_{\leq t_2}$. The basic principle is that if there is a row $r_1 =(s_1,w_1,p_1)$ and a row $r_2 = (s_2,w_2,p_2)$ from the two children, then if each row represents the sum of probabilities of $\mathcal{X}_{r_1}$ and $\mathcal{X}_{r_2}$, respectively, then we can merge these two whenever $s_1=s_2$ (the subframeworks in $\mathcal{X}_{r_1}$ and $\mathcal{X}_{r_2}$ are compatible) and (i) compute $w=w_1\cup w_2$ (we find witnesses for an argument being out or undecided from the respective subframeworks in at least one branch) and (ii) construct $p$ by $\mathtt{comp}(p_1,p_1,s)$, with $s$ containing $F=(A,R,P)$. 
Define 
\begin{align*}
\mathit{common}(s) =\ &\prod_{a \in A}P(a)\cdot \prod_{a \in B_t \setminus A }(1- P(a)) \cdot \\ & \prod_{r \in R} P(r) \cdot \prod_{(a,b) \in D\setminus R} (1 - P(r))\\
\mathtt{comp}(p_1,p_2,s) =\ & \frac{p_1 \cdot p_2}{\mathit{common}(s)}
\end{align*}
with $D$ attacks originally existing and possible in the current bag.
That is, $p$ is computed by the product of $p_1$ and $p_2$, but we have to discount the common part (otherwise the part of $F$ that is the current bag would be counted twice). In case two rows with same $s$ and $w$ are created, $\mathtt{merge}$ merges these into one with summed probabilities.

\begin{algorithm}
\caption{Join($\tau_1,\tau_2,B$)}
\begin{algorithmic}[1]

\STATE{$\tau := \{(s,w_1\cup w_2,\mathtt{comp}(p_1,p_2,s)) \mid (s,w_1,p_1) \in \tau_1, (s,w_2,p_2) \in \tau_2 \}$}

\STATE{$\tau := \mathtt{Merge(\tau)}$}

\STATE{\algorithmicreturn\ $\tau$}

\end{algorithmic}
\end{algorithm}

\section{Experimental Evaluation}

We evaluate our algorithm for $\pext$, focusing on the $\FP^{\#\P}$-complete variant for complete semantics. 

\paragraph{Implementations} 
We implemented our algorithm in a Python (3.9.7) prototype. 
As suggested by~\citeauthor{DewoprabowoFGH22}~(\citeyear{DewoprabowoFGH22}), we partially make use of database tools: the most expensive operations (join and forget) make use of a Python library for database management. 

Due to operations on small numbers, we implemented two variants, one using floating-point numbers dubbed \texttt{TD-Ext-f}, and one using rational numbers, called \texttt{TD-Ext-r}. For computing a nice tree-decompositions, we used the \texttt{htd} library \cite{AbseherMW17}. 

To the best of our knowledge, there are no competing systems available that are tailored towards performance. For a baseline comparison, we implemented two more systems: (i) a naive brute-force algorithm enumerating pairs of subframeworks and extensions in Python and (ii) and answer set programming (ASP)~\cite{GelfondL88,Niemela99} based implementation \texttt{ASP-\#Ext} for \emph{counting} subframeworks where $S$ is complete. Note that this implementation does not solve the same problem, but a simpler one, and is intended mainly as a basis for comparison.

\paragraph{Instances}  

There is currently no standard benchmark set for PAFs. We constructed PAFs with a controlled tree-width and focused on instances not solved via preprocessing, i.e., do not contain many certain arguments and attacks. We constructed PAFs based on grids $(k,n)$ with possible attacks (bidirectional, one-directional, or no attack) for each horizontal or vertical neighbour. Direction or no attacks was chosen with uniform probability. Argument and attack probabilities are picked from $\{0.1, 0.2, ..., 1\}$, with probabilities $\frac{10}{91}$ for uncertain and $\frac{1}{91}$ for $1$. We let $k \in \{3, 4, 5, 6, 7\}$ and $n \in\{5, 10, 20, 50, 75, 100, 150\}$. Tree-width of such grid-graphs is bounded by $min(k, n)$. For each $(k, n)$ we generated 4 PAFs and a set $S$ to be checked by selecting $a \in A$ with probability $0.04$. %

\paragraph{Setup and Results}

The experiments were carried out on a Linux machine (64-bit), with an 8-core i5 Intel CPU and 16 GB of memory. We enforced a runtime of maximum $300$ seconds and a memory limit of $8192$ MB per run. %

As expected, the brute-force algorithm enumerating subframeworks and extensions timed out on most instances. In Table~\ref{tab:median_running_time_complete} we show the number of timeouts and median running times of solved instances. In the results, our approach scales with tree-width: no timeouts were reported with instances having tree-width at most $5$. Note that PAFs with $k=5$ include ones with $n=150$ and $|A| = 750$. For a baseline comparison, \texttt{ASP-\#Ext} showed significantly more timeouts. We speculate this is due to the high number of such subframeworks, e.g., one instance reported more than $6\cdot 10^7$ subframeworks.

Regarding precision, using floats the results, in fact, differ to fractions: in four instances the former reported $0$, while the result is higher than $0$ (although low). Moreover, in one PAF \texttt{TD-Ext-f} reported $9.6\cdot 10^{-15}$ and \texttt{TD-Ext-r} returned $7\cdot 10^{-15}$. While the absolute difference is low, the percentage the former approach is off is roughly one-third. On the other hand, in many runs the results reported were close. We speculate that this is due to having many operations summing or multiplying in the dynamic programming algorithms. Due to similar performance of \texttt{TD-Ext-f} and \texttt{TD-Ext-r}, it does appear to give \texttt{TD-Ext-r} an edge. %

\begin{table}
    \caption{Median running time and timeouts for complete. %
    \label{tab:median_running_time_complete}
}
    \centering
    \scalebox{0.8}{
    \begin{tabular}{|l|l|l|l|}
    \hline
    $k$ & \texttt{TD-Ext-f} & \texttt{TD-Ext-r} & \texttt{ASP-\#Ext} \\ \hline
    3 & 10.49 (0) & 11.02 (0) & 0.01 (21)\\ \hline
  
    4 & 22.51 (0) & 24.09 (0) & 0.20  (22)\\ \hline
  
    5 &  66.43 (0) & 71.76  (0) & 0.37 (23)\\ \hline
 
    6 &  17.38 (8) &  13.53 (9) & 9.61 (25)\\ \hline
 
    7 & 32.76 (10) & 33.74 (11) & 1.33 (24)\\ \hline
   
    \end{tabular}}
\end{table}

\section{Extensions}
\label{sec:extensions}

Our algorithm presented in Section~\ref{sec:alg} can be extended, and we discuss here possible extensions.

First, in addition to enforcing arguments to be ``in'' (as in the $\pext$ problem), we can also directly enforce arguments to be $\labelout$ or $\labelund$. %

This is direct in the algorithm: in introduction nodes we can restrict generation of rows to exactly those fitting the chosen criteria. For instance, only constructing labelings with a chosen argument assigned ``out'' or ``undecided''.

Second, as discussed by~\citeauthor{FazzingaFF19}~(\citeyear{FazzingaFF19}), we can introduce dependencies between arguments. For instance, consider the case that whenever a subframework contains an argument $a$ then such a subframework also must contain an argument $b$. Then, as discussed by~\citeauthor{FazzingaFF19}, one can specify marginal probabilities of the three different cases: containing neither argument, containing only $b$ and containing both. 

This can be achieved, by considering, in the tree-decomposition generation, an additional edge between $a$ and $b$, representing an additional dependency. 
This additional edge ensures that $a$ or $b$ can only be forgotten if the other argument was already seen. %

Then computation of probabilities must be deferred until both arguments $a$ and $b$ are in a bag (only then can we compute the marginal probability of this case). 
As discussed earlier~\cite{Wallner20}, implication dependencies such as these can be important for representing internal structure of arguments, while at the same time, as discussed by~\citeauthor{FazzingaFF19}, allowing for compact representation of probabilities by marginal probabilities. For instance, if an argument $a$ is a sub-argument of $b$, when considering the internal structure of these arguments, then having $b$ but not $a$ might be unwarranted. Or, for instance, if the contents of arguments $a$ and $b$ together imply existence of an argument $c$, then dependency edges between all can be added, and marginal probabilities of the different cases considered.

Finally, to adapt Algorithm~\ref{alg:main} to admissible sets or stable extensions, the condition for semantics can be directly adapted: for admissibility the checks for undecidedness can be omitted, while for stable semantics, introduction nodes are not allowed to assign arguments to undecided. 

\section{Discussion}

In this work we revisited computationally complex problems in probabilistic argumentation under the constellation approach. We refined earlier complexity results, showing a divergence between complexities of two main reasoning tasks. For the problem of computing the probability of a set of arguments being complete, we developed a dynamic progranming algorithm. Our experimental evaluation shows promise of the approach: PAFs with up to 750 arguments were solved regularly, depending on the attack-structure and tree-width.

Algorithmic approaches utilizing tree-decompositions were investigated before for formal argumentation~\cite{Dunne07,DvorakH0SSW22,DvorakMNW11,DvorakPW12,DvorakSW12,FichteH0M21,LampisMM18,PopescuW23}, also for counting extensions containing queried arguments in AFs~\cite{FichteHM19}, and, e.g., for model counting in Boolean logic~\cite{SamerS10} and weighted model counting~\cite{DBLP:conf/esa/FichteHWZ18}. In contrast, we present a novel dynamic programming algorithm for probabilistic argumentation under the constellation approach.

For future work, an interesting avenue is to extend our work for acceptability of arguments in PAFs, under the constellation approach. We think that our Algorithm~\ref{alg:main} provides a useful basis for computing acceptability. The main barrier is to avoid over-computation, or over-summation. For instance, by slightly relaxing our algorithm to have a queried argument to be assigned ``in'', and allowing also other arguments to be ``in'' we can extend the algorithm to consider also other extensions that contain a queried argument. However, directly counting all pairs of subframeworks and labelings assigning a queried argument to be ``in'' over-counts, since there might be many such labelings in a particular subframework. 

We think that this way of over-counting can be addressed in a dynamic programming way operating on tree-decompositions, but requires more work for a detailed investigation. 
We remark that Theorem~\ref{thm:complex-acc-acyclic} (hardness of acyclic PAFs) does not prevent existence of algorithms for acceptability using tree-decompositions: acyclic PAFs may have a tree-width higher than one. %
  
As a more general direction, we think that future work, to advance complex reasoning in probabilistic argumentation further, should include also different forms of optimizations. For instance, an ($\NP$-hard) preprocessing might be beneficial for dynamic programming algorithms that can restrict the number of rows in tables, e.g, by pre-computing impossible combinations (e.g., argument $a$ is never ``in'' in any subframework). While these problems can be complex themselves, we think that they can overall be used to optimize tree-decomposition-based algorithms.

\bibliographystyle{kr}
\bibliography{library}

\end{document}